\documentclass[letterpaper]{article} 
\usepackage{aaai25}  
\usepackage{times}  
\usepackage{helvet}  
\usepackage{courier}  
\usepackage[hyphens]{url}  
\usepackage{graphicx} 
\usepackage{natbib}  
\usepackage{caption} 
\usepackage{algorithm}
\usepackage{algorithmic}

\usepackage{bm} 
\usepackage{amsmath, amssymb, amsthm, epsfig, url, array}
\usepackage{cleveref}

\theoremstyle{plain}
\newtheorem{theorem}{Theorem}[section]

\newtheorem{proposition}[theorem]{Proposition}
\newtheorem{lemma}[theorem]{Lemma}

\theoremstyle{definition}
\newtheorem{definition}[theorem]{Definition}
\newtheorem{assumption}[theorem]{Assumption}
\newtheorem{example}[theorem]{Example}
\crefname{theorem}{theorem}{theorems}
\crefname{lemma}{lemma}{lemmas}
\crefname{definition}{definition}{definitions}
\crefname{assumption}{assumption}{assumptions}
\crefname{corollary}{cororally}{corollaries}
\crefname{property}{property}{properties} 
\usepackage {threeparttable}
\usepackage{makecell, caption, booktabs, multirow, siunitx}

\usepackage{newfloat}
\usepackage{listings}
\DeclareCaptionStyle{ruled}{labelfont=normalfont,labelsep=colon,strut=off} 
\floatstyle{ruled}
\newfloat{listing}{tb}{lst}{}
\floatname{listing}{Listing}
\title{Decomposed Quadratization: \\ Efficient QUBO Formulation for Learning Bayesian Network}
\author{
    Yuta Shikuri
}
\affiliations{
    Tokio Marine Holdings, Inc. \\ 
    Tokyo, Japan  \\ 
    shikuriyuta@gmail.com
}

\begin{document}

\maketitle

\begin{abstract}
Algorithms and hardware for solving quadratic unconstrained binary optimization (QUBO) problems have made significant recent progress. 
This advancement has focused attention on formulating combinatorial optimization problems as quadratic polynomials. 
To improve the performance of solving large QUBO problems, it is essential to minimize the number of binary variables used in the objective function. 
In this paper, we propose a QUBO formulation that offers a bit capacity advantage over conventional quadratization techniques. 
As a key application, this formulation significantly reduces the number of binary variables required for score-based Bayesian network structure learning. 
Experimental results on $16$ instances, ranging from $37$ to $223$ variables, demonstrate that our approach requires notably fewer binary variables than quadratization.  
Moreover, an annealing machine that implement our formulation have outperformed existing algorithms in score maximization. 
\end{abstract}

\begin{figure*}[t]
  \begin{center}
    \includegraphics[width=175mm]{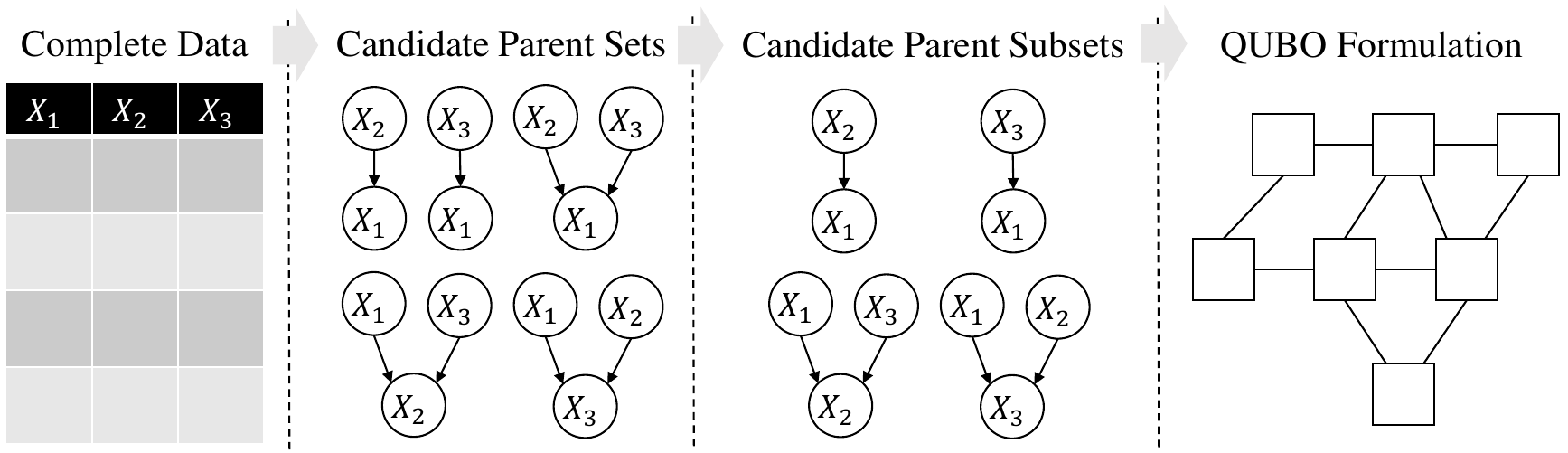}
  \end{center}
\caption{
  Overview of our approach to Bayesian network structure learning. 
  We introduce the concept of candidate parent subsets to represent candidate parent sets as a quadratic polynomial.     
}
\label{overview}
\end{figure*}

\section{Introduction}
A Bayesian network is a probabilistic graphical model that represents a joint probability distribution among random variables in a directed acyclic graph (DAG) \citep{Judea}. 
One class of associated computational problems is learning the graph structure of a Bayesian network from data. 
There are two principal approaches to Bayesian network structure learning: constraint-based and score-based \citep{Neville}. 
Constraint-based algorithms construct graphs through conditional independence tests. 
Score-based approaches aim to find a DAG with the highest possible score. 
In this study, we focus on score-based approaches. 

Bayesian network structure learning is NP-hard \citep{David}. 
Many approaches have been proposed to improve accuracy and reduce execution time. 
For a small number of variables, some algorithms such as \citep{Cussens} are capable of identifying a DAG with the maximum score. 
For high-dimensional data, the standard methodology involves the use of approximate approaches. 
The hill climbing search method over the space of DAGs \citep{Remco} remains competitive despite its simplicity. 
Approximate searches over the space of topological ordering \citep{Marc} are also well known as competitive algorithms. 

Quadratic unconstrained binary optimization (QUBO) is capable of modeling a wide range of combinatorial optimization problems, 
including image synthesis in the field of computer vision \citep{Vladimir} and various machine learning tasks such as neural networks \citep{Michele} and decision trees \citep{Koichiro}.  
QUBO can also be applied to score-based Bayesian network structure learning \citep{Bryan}.  
Note that constraint-based algorithms do not appear to be suitable for QUBO formulation. 
Recent developments in annealing machines have garnered significant attention for QUBO formulation. 
An annealing machine is a specialized hardware architecture designed to heuristically solve QUBOs \citep{Kasho}. 
In particular, annealing machines based on semiconductor technology derived from simulated annealing have reportedly outperformed existing solvers in maximum cut problems \citep{Satoshi}. 

The challenge of setting up a QUBO lies in reformulating the objective function into a quadratic polynomial using fewer binary variables. 
The greater the number of binary variables in a QUBO, the more difficult it generally becomes to solve. 
More practically, encoding a large number of binary variables onto the hardware circuit of an annealing machine is often infeasible. 
Quadratization is the common method of QUBO formulation \citep{Elisabeth, Martin, Endre}. 
Algorithms have been developed to minimize the number of auxiliary variables for quadratization \citep{Amit}.  
However, quadratization still requires too many auxiliary variables for certain tasks, including Bayesian network structure learning.  
Consequently, there is a potential demand for a more efficient QUBO formulation method as an alternative to conventional quadratization. 

In this study, we propose a QUBO formulation specifically designed to use fewer binary variables than quadratization.  
We demonstrate that this formulation offers significant advantages for learning Bayesian networks in terms of bit capacity.  
\Cref{overview} illustrates the overview of our approach. 
Experimental results on $16$ instances, ranging from $37$ to $223$ variables, show that our formulation dramatically reduces the number of binary variables compared to \citep{Bryan}. 
These instances can be encoded on the circuit of the Fujitsu Digital Annealer, a fully-coupled annealing machine with $100$K bit capacity \citep{Hiroshi}. 
The scores achieved by the Digital Annealer using our formulation were greater than those of existing approaches.

\section{Preliminary} 
We describe a QUBO formulation for learning Bayesian networks, incorporating candidate parent sets into the approach of \citep{Bryan}.  
For indexed symbols, we assign the zero index to the empty set and ensure no overlap (e.g., $W_{i0} = \emptyset$ and $\{W_{ij}\}_{j = 0}^{\lambda_i}$ includes $\lambda_i + 1$ elements).  
\Cref{Description} provides the list of symbols.

\subsection{Bayesian Network Structure Learning}
A Bayesian network, which is a graphical model composed of a DAG and its parameters, represents a joint probability distribution.  
Each vertex of the DAG corresponds to a random variable. 
Edges and parameters characterize conditional probability distributions. 
Let $\mathcal{X} \equiv \{X_i\}_{i=1}^n$ denote the set of vertices, and $\Pi_i$ denote the set of vertices with edges directed towards $X_i$. 
A topological ordering of a graph exists if and only if the graph is a DAG \citep{Thomas}. 
\begin{definition}
  \label{to}
  A topological ordering $\prec$ is a binary relation between any two of vertices such that 
  \begin{itemize}
    \item $X_i \prec X_j$ and $X_j \prec X_k \Rightarrow X_i \prec X_k$.  
    \item $X_i \prec X_j \Rightarrow X_i \notin \Pi_j$.  
  \end{itemize}
\end{definition}
The goal of score-based Bayesian network structure learning is to find a DAG with the maximum score. 
A score is the sum of local scores which only rely on the parent set of each vertex. 
Given complete data, the parent sets $\Pi_1, \cdots, \Pi_n$ are optimized to maximize the score under the constraint that the graph $\mathcal{G}$ corresponding to them is a DAG. 
This optimization is formulated as  
\begin{align} 
  \label{BNSL}
  \mathrm{maximize~} \sum_{1 \leq i \leq n} \log S_i (\Pi_i) \mathrm{~~subject~to~} \mathcal{G} \mathrm{~is~a~DAG},  
\end{align} 
where $S_i : 2^{\mathcal{X} \setminus \{X_i\}} \rightarrow \mathbb{R}$. 
For simplicity, we take $S_i (W) = S_i (\emptyset)$ when $|W|$ is greater than the maximum parent set size $m$.  
The identification of candidate parent sets facilitates narrowing the search space by the relation between parent sets and local scores \citep{Qiang}. 
\begin{definition} 
  \label{CPS}
  The candidate parent sets of $X_i$ are defined as $\{W \subseteq \mathcal{X} \setminus \{X_i\} \mid W^\prime \subset W \Rightarrow S_i (W^\prime) < S_i (W)\}$.  
\end{definition} 
Let $\{W_{ij}\}_{j = 0}^{\lambda_i}$ denote the candidate parent sets of $X_i$, and $\mathcal{X}_i$ denote the union of them. 
Using the candidate parent sets, we can ignore the topological ordering of certain edges that do not produce cycles. 
Let $\mathcal{E}$ denote the set of edges on possible cycles.  

\subsection{QUBO Formulation} 
QUBO is a mathematical optimization problem of minimizing a quadratic polynomial with binary variables. 
A pseudo-Boolean function maps binary-valued inputs to a real value. 
Every pseudo-Boolean function can be uniquely represented as a multilinear polynomial given by  
\begin{align} 
  f(\bm{p}) \equiv \sum_{V \in \mathcal{F}} \pi (V) \prod_{i \in V} p_i, 
\end{align} 
where $\bm{p} \equiv (p_i)_{i=1}^I \in \{0, 1\}^I$, $\pi : 2^{\{1, \cdots, I\}} \rightarrow \mathbb{R} \setminus \{0\}, \mathcal{F} \subseteq 2^{\{1, \cdots, I\}} \setminus \{\emptyset\}$, and we ignore the constant term.  
Quadratization is a major technique to convert a higher degree pseudo-Boolean function into a quadratic one using auxiliary variables $\bm{q} \equiv (q_i)_{i=1}^J \in \{0, 1\}^J$.  
\begin{definition} 
The quadratization of $f$ is a quadratic polynomial function $g : \{0, 1\}^I \times \{0, 1\}^J \rightarrow \mathbb{R}$ such that 
\begin{align}
  f(\bm{p}) = \min_{\bm{q} \in \{0, 1\}^J} g(\bm{p}, \bm{q}). 
\end{align}   
\end{definition} 
\citeauthor{Rosenberg} \citeyear{Rosenberg} has proven that every pseudo-Boolean function can be transformed into a quadratic polynomial through the substitution of the product of two variables with an auxiliary variable and the addition of a penalty term.

\subsection{QUBO Formulation for Structure Learning} 
\label{BasicQUBO} 
Bayesian network learning can be uniquely represented by a multilinear polynomial over the state of edges and topological ordering. 
The state of $\bm{d} \equiv ((d_{ij})_{j = 1}^n)_{i = 1}^n$ is mapped one-to-one to the set of edges, where $d_{ij} = 1$ if $X_j$ is a parent of $X_i$; otherwise, $d_{ij} = 0$.   
The score component is 
\begin{align} 
    O (\bm{d}) \equiv \sum_{1 \leq i \leq n} \sum_{W \subseteq \mathcal{X}_i} \pi_i(W) \prod_{1 \leq j \leq n; X_j \in W} d_{ij}, 
\end{align} 
where $\pi_i(W) \equiv \sum_{W^\prime \subseteq W}(-1)^{1 + |W \setminus W^\prime|}\log S_i (W^\prime)$. 
The higher-degree terms in the score component can be transformed into quadratic terms through quadratization.  
The state of $\bm{r} \equiv (r_{ij})_{(i, j) \in \mathcal{E}}$ corresponds to the topological ordering, where $r_{ij} = 0$ if the order of $X_j$ is higher than $X_i$; otherwise, $r_{ij} = 1$. 
The linear ordering of vertices and the consistency of edges  in \cref{to} are captured by  
\begin{align} 
    \label{Hcycle}
    C (\bm{d}, \bm{r}) \equiv &\sum_{(i, j), (j, k), (i, k) \in \mathcal{E}} \delta_1 R(r_{ij}, r_{jk}, r_{ik})& \nonumber  \\   
    &+ \sum_{(i, j) \in \mathcal{E}} \delta_2 \bigl(d_{ij} r_{ij} + d_{ji} (1 -r_{ij})\bigr),&  
\end{align} 
where $\delta_1, \delta_2 \in (0, \infty)$ and $R(r_{ij}, r_{jk}, r_{ik}) \equiv r_{ik} + r_{ij} r_{jk} - r_{ij} r_{ik} - r_{jk} r_{ik}$.  
There exist $(\delta_1, \delta_2)$ such that $C (\bm{d}, \bm{r}) = 0$ when the objective function $O (\bm{d}) + C (\bm{d}, \bm{r})$ is minimized. 
Additionally, the graph corresponding to the state of $\bm{d}$ is a DAG if $C (\bm{d}, \bm{r}) = 0$.   
Consequently, this QUBO formulation is equivalent to learning Bayesian networks.

\section{Decomposed Quadratization}
\label{DQ} 
Searching over candidate parent sets is a key technique in Bayesian network structure learning. 
To deal with candidate parent sets in a QUBO formulation, we use auxiliary variables corresponding to the sets of primary variables. 
The auxiliary variables capture the score component and identify the state of primary variables.  
We decompose a quadratization into the optimization of auxiliary variables and the transformation that maps them to primary variables. 
This decompostion of quadratization is defined as 
\begin{definition} 
    \label{decomposition}
    The decomposed quadratization of $f$ is the pair of a function $g_1 : \{0, 1\}^J \rightarrow \{0, 1\}^I$ and a quadratic polynomial function $g_2 : \{0, 1\}^J \rightarrow \mathbb{R}$    
    such that  
    \begin{align}
      f(\bm{p}) = \min_{\bm{q} \in \{0, 1\}^J; \bm{p} = g_1(\bm{q})} g_2(\bm{q}).  
    \end{align}   
\end{definition} 
\begin{example}
  Let $g_1(q_1, q_2) = (q_1, q_2, q_2)$ and $g_2(q_1, q_2) = - q_1 - 3 q_2 + 2 q_1 q_2$. 
  Then $(g_1, g_2)$ is a decomposed quadratization of $f(p_1, p_2, p_3) = - p_1 - 3 p_2 p_3 + 2 p_1 p_2 p_3$. 
\end{example}  
Here we describe the behavior of decomposed quadratization. 
Considering the case where each $p_i$ corresponds to $q_i$, the concept of decomposed quadratization encompasses quadratization. 
To capture the subsets of the terms in $\mathcal{F}$, we make a set $\emptyset \subset Q_i \subseteq \{1, \cdots, I\}$ correspond to a binary variable $q_i \in \{0, 1\}$ one-to-one. 
Let $\bar{g}_1, \bar{g}_2$ denote  
\begin{align} 
  \label{QUBO_DQ}
  \bar{g}_1 (\bm{q}) \equiv (\min \{1, \sum_{1 \leq j \leq J; i \in Q_j} q_j\})_{i=1}^I, \\ 
  \bar{g}_2 (\bm{q}) \equiv \sum_{1 \leq i \leq j \leq J} \bar{\pi} (Q_i, Q_j) q_i q_j + h(\bm{q}), 
\end{align} 
where $\bar{\pi} : 2^{\{1, \cdots, I\}} \times 2^{\{1, \cdots, I\}} \rightarrow \mathbb{R}$, and $h : \{0, 1\}^J \rightarrow [0, \infty)$is a quadratic polynomial. 
To discuss the behavior of $(\bar{g}_1, \bar{g}_2)$, we introduce \cref{theorem_decomposition}. 
\Cref{r1} ensures the comprehensiveness of the terms in $\mathcal{F}$. 
\Cref{r2} presents the inclusion relations among $\{Q_i\}_{i=1}^J$. 
\begin{assumption}  
  \label{r1} 
  $\mathcal{F} \subseteq \bigcup_{1 \leq i \leq j \leq J} \{Q_i \cup Q_j\}$.   
\end{assumption}  
\begin{assumption}
  \label{r2} 
  If there exists a pair of an index $1 \leq i \leq J$ and a set $\mathcal{W} \subseteq \{1, \cdots, J\} \setminus \{i\}$ with $|\mathcal{W}| \geq 2$ such that  
  \begin{align} 
    Q_i \subseteq \bigcup_{j \in \mathcal{W}} Q_j \mathrm{~and~} Q_i \nsubseteq \bigcup_{j \in \mathcal{W}^\prime} Q_j \mathrm{~for~any~} \mathcal{W}^\prime \subset \mathcal{W},  
  \end{align}   
  then $Q_i = Q_j \cup Q_k$ for some $j, k \in \{1, \cdots, J\} \setminus \{i\}$. 
\end{assumption} 
\begin{theorem} 
    \label{theorem_decomposition}
    Suppose that \cref{r1} and \cref{r2} hold. 
    Then there exists $(\bar{\pi}, h)$ such that $(\bar{g}_1, \bar{g}_2)$ is a decomposed quadratization of $f$. 
\end{theorem} 
\begin{proof}
    See \cref{proof_dq}.     
\end{proof}
To capture the inclusion relations, we consider designing a penalty function $h$ that induces 
\begin{align} 
  \label{cover} 
  Q_i \subseteq \bigcup_{j \in \mathcal{W}} Q_j \Rightarrow (1 - q_i) \prod_{j \in \mathcal{W}} q_j = 0. 
\end{align} 
According to \cref{exist0}, one element cannot be directly covered by more than three others. 
This prevents the direct incorporation of \cref{cover} into a penalty. 
\Cref{r2} indirectly addresses it by covering one element with two others. 
\Cref{exist1} restricts the formulation of penalties to capture this assumption. 
\begin{lemma} 
    \label{exist0} 
    For $K \in \{3, 4, \cdots \}$, no quadratic polynomial function $\phi : \{0, 1\}^{K + 1} \rightarrow \mathbb{R}$ satisfies  
    \begin{align} 
      \label{qqq}
      \phi(q_1, \cdots, q_K, 0) = 0 \Leftrightarrow \prod_{1 \leq i \leq K} q_i = 0.   
    \end{align}     
\end{lemma} 
\begin{proof}
    See \cref{proof_exist0}. 
\end{proof}
\begin{lemma} 
  \label{exist1} 
  Suppose that a quadratic polynomial function $\phi : \{0, 1\}^3 \rightarrow \mathbb{R}$ satisfies \cref{qqq}. 
  Then the following holds: 
  \begin{align} 
    \phi(0, 0, 1) + \phi(1, 1, 1) \neq \phi(1, 0, 1) + \phi(0, 1, 1). 
  \end{align}     
\end{lemma} 
\begin{proof}
  See \cref{proof_exist1}. 
\end{proof}
Quadratization is a specific form of decomposed quadratization that requires \cref{R_qua} in addition to \cref{r1} and \cref{r2}. 
This additional assumption of quadratization is disadvantageous for reducing the number of auxiliary variables in certain tasks such as \cref{example_DQ}. 
\begin{assumption}  
    \label{R_qua} 
    $\{i\} \in \{Q_j\}_{j=1}^J$ for all $1 \leq i \leq I$. 
\end{assumption}  
\begin{example}
  \label{example_DQ}
  Let $\mathcal{F} = \{\{1, 2, 3, 4\}, \{1, 2, 3\}, \{4\}, \{3, 4\}\}$.  
  A decomposed quadratization requires $3$ auxiliary variables: $Q_1 = \{1, 2, 3\}, Q_2 = \{4\}$, and $Q_3 = \{3\}$.  
  A quadratization requires $6$ auxiliary variables: $Q_1 = \{1\}, Q_2 = \{2\}, Q_3 = \{3\}, Q_4 = \{4\}, Q_5 = \{1, 2\}$, and $Q_6 = \{3, 4\}$. 
\end{example}
Considering the biases and variable couplers in a QUBO, the auxiliary variables satisfy $J \geq \lceil - \frac{1}{2} + \frac{1}{2} \sqrt{1 + 8 |\mathcal{F}|} \rceil$. 
We minimize the number of auxiliary variables to capture \cref{r1} and \cref{r2}. 
An optimal solution of $\{Q_i\}_{i=1}^J$ is a subset of $\bigcup_{V \in \mathcal{F}} 2^V \setminus \{\emptyset\}$. 
Let $V_\mathrm{close}, V_\mathrm{open} \subseteq \bigcup_{V \in \mathcal{F}} 2^V \setminus \{\emptyset\}$ 
such that $V_\mathrm{close} \cap V_\mathrm{open} = \emptyset$, $V_\mathrm{close}$ is a subset of an optimal solution, and $V_\mathrm{close} \cup V_\mathrm{open}$ contains an optimal solution.  
Since $V_\mathrm{close}$ is a subset of an optimal solution, we can ignore the terms in $\mathcal{F}$ that can be represent by the union of some two elements in $V_\mathrm{close}$. 
Let $W_\mathrm{open} \subseteq \mathcal{F}$ be the terms that still cannot be ignored. 
We provide \cref{SSR0} to find an optimal solution. 
The mechanism of this algorithm is as follows: 
\begin{description}
  \item[Row 2:] If there is only one way to represent $W$, then $V$ is included in an optimal solution. 
  \item[Row 3:] We can ignore elements in $W_\mathrm{open}$ represented by pairs of elements in $V_\mathrm{close}$. 
  \item[Row 4:] If $V$ is not a subset of any element in $W_\mathrm{open}$, then it is not included in an optimal solution. 
\end{description} 

Here we minimize $\sum_{i=1}^{\nu} v_i$ under the constraint that \cref{r1} and \cref{r2} hold for  
\begin{align} 
  \{Q_i\}_{i = 1}^J = V_\mathrm{close} \cup \{V_i \mid v_i = 1\}_{i = 1}^\nu, 
\end{align} 
where $\bm{v} \equiv (v_i)_{i=1}^{\nu} \in \{0,1\}^{\nu}$ and $\{V_i\}_{i=1}^{\nu} \equiv V_\mathrm{open}$.   
By expressing the constraints as linear inequalities, the search for an optimal solution can be formulated as an integer linear programming problem of $\bm{v}$. 
See \cref{Optimization}. 
This formulation, with some conversion, can be adapted to minimize the number of terms. 
See \cref{term}. 

\begin{algorithm}[t] 
  \caption{Search Space Reduction} 
  \label{SSR0} 
  \begin{algorithmic}[1] 
  \STATE $W_\mathrm{open} \leftarrow \mathcal{F}, V_\mathrm{close} \leftarrow \emptyset, V_\mathrm{open} \leftarrow \bigcup_{V \in \mathcal{F}} 2^V \setminus \{\emptyset\}$. 
  \STATE Merge $\{V \in V_\mathrm{open} \mid$ for some $W \in W_\mathrm{open}$, $V$ is only an element in $V_\mathrm{open}$ such that $W \in \bigcup_{V^\prime \in \{\emptyset\} \cup V_\mathrm{close}} \{V \cup V^\prime\}\}$ into $V_\mathrm{close}$.  
  \STATE Exclude $\bigcup_{V, V^\prime \in V_\mathrm{close}} \{V \cup V^\prime\}$ from $W_\mathrm{open}$.  
  \STATE Exclude $V_\mathrm{close} \cup \{V \in V_\mathrm{open} \mid$ there does not exist $W \in W_\mathrm{open}$ such that $V \subseteq W\}$ from $V_\mathrm{open}$. 
  \end{algorithmic}
\end{algorithm}

\begin{figure*}[t]
  \begin{center}
  \includegraphics[width=170mm]{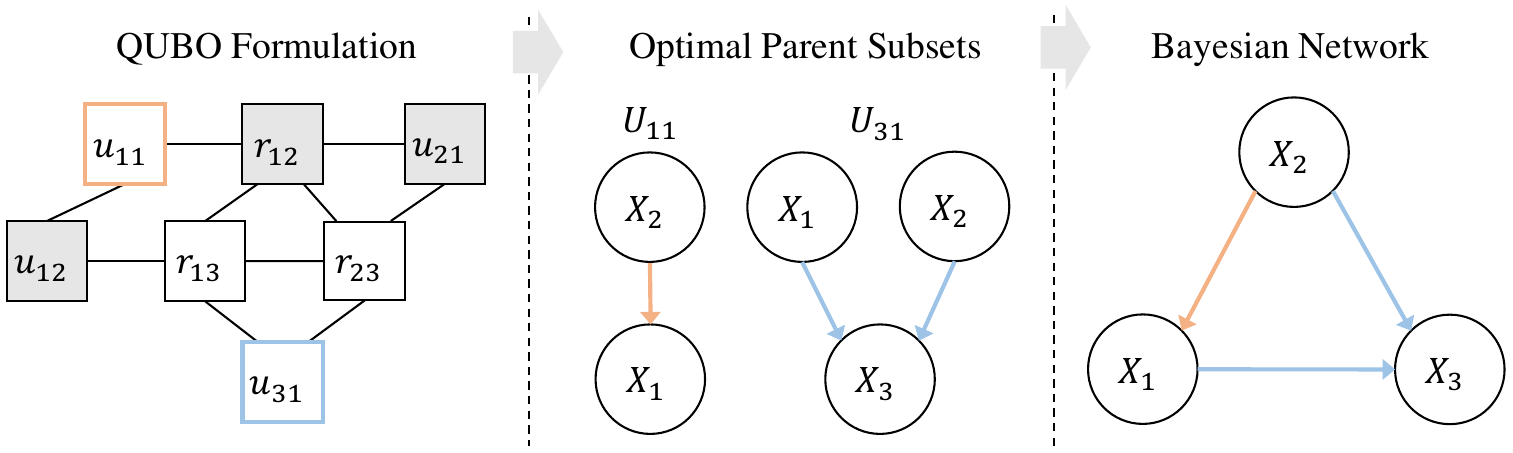}
  \end{center}
  \caption{
  Example of the QUBO formulation with candidate parent subsets. 
  An undirected edge between two binary variables represents that the objective function contains their product term.  
  Gray signifies that the state of a binary variable is $0$ and white is $1$. 
  Each color corresponds to the edge in the DAG.  
  Let $n = 3, m = 2, \lambda_1 = 3, \lambda_2 = 1, \lambda_3 = 1, \Pi_1 = \{X_2\}, \Pi_2 = \emptyset, \Pi_3 = \{X_1, X_2\}, W_{11} = \{X_2\}, W_{12} = \{X_3\}, W_{13} = \{X_2, X_3\}, W_{21} = \{X_1, X_3\}, W_{31} = \{X_1, X_2\}, \mu_1 = 2, \mu_2 = 1, \mu_3 = 1, U_{11} = \{X_2\}, U_{12} = \{X_3\}, U_{21} = \{X_1, X_3\}$, and $U_{31} = \{X_1, X_2\}$. 
  }
  \label{EC} 
\end{figure*}

\section{Efficient QUBO for Structure Learning} 
\label{QUBO_Decomposition}
In this section, we propose the decomposed quadratization of $O$ to encode candidate parent sets efficiently. 
Here we introduce the concept of candidate parent subsets. 
\begin{definition} 
  \label{candidate_parent_subsets} 
  The candidate parent subsets of $X_i$ are defined as the set containing elements $U, U^\prime \subseteq \mathcal{X}_i$ such that $U \cup U^\prime = W$, for any $W$ in the candidate parent sets of $X_i$. 
\end{definition}   
Let $\{U_{ij}\}_{j = 0}^{\mu_i}$ denote the candidate parent subsets of $X_i$. 
To encode candidate parent subsets in a QUBO, we make a candidate parent subset $U_{ij}$ correspond to a binary variable $u_{ij} \in \{0, 1\}$ one-to-one.  
The state of $\bm{d}$ is identified by  
\begin{align} 
  D(\bm{u}) \equiv ((\min\{1, \sum_{1 \leq k \leq \mu_i; X_j \in U_{ik}} u_{ik}\})_{j = 1}^n)_{i = 1}^n,  
\end{align}  
where $\bm{u} \equiv ((u_{ij})_{j = 1}^{\mu_i})_{i = 1}^n$. 
A parent set is captured by the union of at most two elements included in candidate parent subsets. 
Since each vertex has just one parent set, we can disregard the inclusion relations among candidate parent subsets as mentioned in \cref{r2}. 
From \cref{candidate_parent_subsets}, the following proposition equivalent to \cref{r1} holds. 
\begin{proposition}  
  \label{r3} 
  $\mathcal{F}_i \equiv \{W_{ij}\}_{j = 1}^{\lambda_i} \subseteq \bigcup_{1 \leq j \leq k \leq \mu_i} \{U_{ij} \cup U_{ik}\} \subseteq 2^{\mathcal{X}_i} \setminus \{\emptyset\}$ for all $1 \leq i \leq n$.   
\end{proposition}  
The main distinction of our approach from \cref{BasicQUBO} is that the objective function is represented not by edges among vertices but by candidate parent subsets.   
To enforce exactly one parent set per vertex (i.e., $\sum_{1 \leq j \leq \mu_i} u_{ij} \leq 2$), we use the constraint characterized by a penalty coefficient $\xi \in (0, \infty)$ and auxiliary variables $\bm{z} = (z_i)_{i = 1}^n \in \{0, 1\}^n$.    
Then the score component is 
\begin{align} 
    \bar{O}(\bm{u}, \bm{z}) \equiv \sum_{1 \leq i \leq n} \sum_{1 \leq j \leq k \leq \mu_i} \bar{\pi}_i (U_{ij}, U_{ik}) u_{ij} u_{ik}    \nonumber  \\ 
    + \sum_{1 \leq i \leq n} \xi \Bigl(z_i - z_i \sum_{1 \leq j \leq \mu_i} u_{ij} + \sum_{1 \leq j < k \leq \mu_i} u_{ij} u_{ik}\Bigr),   
\end{align} 
where $\bar{\pi}_i (U, U^\prime) \equiv (-1)^{1 + |\{U\} \cup \{U^\prime\}|} (\log S_i (U \cup U^\prime) - \log S_i (U) - \log S_i (U^\prime) + \log S_i (\emptyset))$. 
Using \cref{r3}, our formulation requires equal to or fewer binary variables than the quadratization of $O$. 
For candidate parent subsets in quadratization, it is essential to capture the relation between them and the edges as follows: 
\begin{align} 
  (1 - u_{ij}) \prod_{1 \leq k \leq n; X_k \in U_{ij}} d_{ik} = 0. 
\end{align}   
\Cref{exist0} demonstrates that such relation cannot be captured by a quadratic polynomial when a candidate parent subset includes more than three elements.  
Consequently, we benefit from the use of decomposed quadratization with candidate parent subsets. 
The following lemma provides a lower bound for the penalty coefficient $\xi$.  
\begin{lemma} 
  \label{sufficient1} 
  Let $\xi_0$ be defined as  
  \begin{align} 
  \xi_0 \equiv - 3 \min_{1 \leq i \leq n} \min_{0 \leq j \leq k \leq \mu_i} \bar{\pi}_i (U_{ij}, U_{ik}).    
  \end{align}  
  Suppose that $\xi > \xi_0$ holds and the state of $\bm{d}$ aligns with candidate parent sets. 
  Then $(D, \bar{O})$ is a decomposed quadratization of $O$. 
\end{lemma} 
\begin{proof} 
    See \cref{Proof_Bounds1}. 
\end{proof} 
The constraint for \cref{to} is also captured by candidate parent subsets. 
Let $\bar{C} (\bm{u}, \bm{r})$ denote the right side of \cref{Hcycle} where $d_{ij}$ is replaced by  
\begin{align} 
  \sum_{1 \leq k \leq \mu_i; X_j \in U_{ik}} u_{ik}, 
\end{align}  
where this substitution does not use additional auxiliary variables, but may increase the number of terms.    
The objective function in our approach is $\bar{H} (\bm{u}, \bm{r}, \bm{z}) \equiv \bar{O} (\bm{u}, \bm{z}) + \bar{C} (\bm{u}, \bm{r})$. 
\Cref{EC} illustrates this formulation. 
The following theorem guarantees that optimal networks can be obtained using our QUBO formulation. 
\begin{theorem} 
\label{sufficient2} 
    Let $\delta_0$ be defined as   
    \begin{align} 
    \delta_0 \equiv \max_{1 \leq i \leq n, 0 \leq j \leq \lambda_i} (\log S_i (W_{ij}) - \log S_i (\emptyset)).    
    \end{align} 
    Suppose that $\xi > \xi_0$ and $\delta_2 > (n - 2) \delta_1 > (n - 2) \delta_0$ hold. 
    Then $\bar{C} (\bm{u}, \bm{r}) = 0$ holds when $\bar{H}(\bm{u}, \bm{r}, \bm{z})$ is minimized. 
\end{theorem} 
\begin{proof} 
    See \cref{Proof_Bounds2}. 
\end{proof}

\begin{algorithm}[t]
  \caption{Candidate Parent Subset Reduction}  
  \label{SSR} 
  \begin{algorithmic}[1]
  \STATE $W_\mathrm{open} \leftarrow \mathcal{F}_i, V_\mathrm{close} \leftarrow \emptyset, V_\mathrm{open} \leftarrow \bigcup_{V \in \mathcal{F}_i} 2^V \setminus \{\emptyset\}$.  
  \WHILE{$V_\mathrm{close}$ has changed since the last evaluation,}  
  \STATE Exclude $\{V \in V_\mathrm{open} \mid V \subseteq W$ for at most one element $W \in W_\mathrm{open}\}$ from $V_\mathrm{open}$. 
  \STATE Execute the row from $2$ to $4$ in \cref{SSR0}. 
  \STATE Merge $W_\mathrm{open} \setminus \bigcup_{V, V^\prime \in V_\mathrm{close} \cup V_\mathrm{open}} \{V \cup V^\prime\}$ into $V_\mathrm{close}$, and exclude it from $W_\mathrm{open}$.  
  \ENDWHILE 
  \STATE Merge $W_\mathrm{open} \setminus \bigcup_{V \in V_\mathrm{open}, V^\prime \in V_\mathrm{close}} \{V \cup V^\prime\}$ into $V_\mathrm{open}$. 
  \end{algorithmic}
\end{algorithm}  

The number of required binary variables is $\sum_{1 \leq i \leq n} (\mu_i + 1) + |\mathcal{E}|$. 
We find candidate parent subsets to minimize $\mu_i$ under the constraint to capture \cref{r3}. 
An optimal solution of candidate parent subsets of $X_i$ is a subset of $\bigcup_{V \in \mathcal{F}_i} 2^V \setminus \{\emptyset\}$. 
Since \cref{r2} is not required, \cref{SSR} can reduce the search space of candidate parent subsets. 
This algorithm works as follows:   
\begin{description}
  \item[Row 3:] We temporarily ignore elements that contribute to represent at most one element included in $W_\mathrm{open}$. 
  \item[Row 5:] Considering the removed elements at row 3, an optimal solution includes the elements in $W_\mathrm{open}$ that cannot be represented by pairs of elements in $V_\mathrm{close} \cup V_\mathrm{open}$. 
  \item[Row 7:] From a perspective similar to row 5, elements in $W_\mathrm{open}$ that are represented solely by pairs of elements in $V_\mathrm{open}$ may be included in an optimal solution.  
\end{description} 
Additionally, we can exclude the constraint that captures \cref{r2} from the integer linear programming problem.

\begin{table*}[t]
  \centering
  \begin{threeparttable}[]
  \begin{tabular}{cccccccc} 
  \toprule   
  \multirow{2.5}{*}{Instance} & \multirow{2.5}{*}{$m$} & \multicolumn{2}{c}{\citeauthor{Bryan}} & \multicolumn{2}{c}{IP only} & \multicolumn{2}{c}{IP + \cref{SSR}} \\ 
  \cmidrule(lr){3-4} \cmidrule(lr){5-6} \cmidrule(lr){7-8} &   & Number & Time [s] &  Number & Time [s] & Number & Time [s]  \\ 
  \midrule 
  alarm & $4$ & $3982 \pm 346$ & $272 \pm 61$ & $\bm{1373 \pm 46}$ & $61 \pm 29$ & $\bm{1373 \pm 46}$ & $\bm{14 \pm 2}$  \\ 
  barley & $4$ & $1826 \pm 279$ & $271 \pm 149$ & $\bm{995 \pm 143}$ & $\bm{11 \pm 2}$ & $\bm{995 \pm 143}$ & $\bm{11 \pm 2}$  \\    
  hailfinder & $4$ & $2374 \pm 31$ & $52 \pm 24$ & $\bm{1913 \pm 6}$ & $\bm{4 \pm 1}$ & $\bm{1913 \pm 6}$ & $\bm{4 \pm 1}$   \\    
  hepar2 & $4$ & $3629 \pm 423$ & $51 \pm 16$ & $\bm{2616 \pm 133}$ & $3 \pm 0$ & $\bm{2616 \pm 133}$ & $\bm{2 \pm 0}$  \\ 
  \midrule 
  chess & $2$ & $7040 \pm 167$ & $\bm{63 \pm 2}$ & $\bm{6964 \pm 165}$ & $89 \pm 4$ & $\bm{6964 \pm 165}$ & $103 \pm 4$   \\ 
  chess & $3$ & $80116 \pm 3014$ & $\bm{3229 \pm 151}$ & $26355 \pm 1403$ & $7268 \pm 492$ & $\bm{23662 \pm 1103}$ & $5567 \pm 376$  \\    
  win95pts & $2$ & $5150 \pm 39$ & $\bm{27 \pm 2}$ & $\bm{5028 \pm 42}$ & $33 \pm 3$ & $\bm{5028 \pm 42}$ & $37 \pm 4$  \\    
  win95pts & $3$ & $37705 \pm 1935$ & $1223 \pm 90$ & $9220 \pm 398$ & $1036 \pm 117$ & $\bm{9025 \pm 333}$ & $\bm{890 \pm 105}$  \\    
  pathfinder & $2$ & $10831 \pm 382$ & $\bm{178 \pm 8}$ & $\bm{10725 \pm 379}$ & $259 \pm 30$ & $\bm{10725 \pm 379}$ & $284 \pm 20$   \\    
  pathfinder & $3$ & $107725 \pm 5329$ & $\bm{11695 \pm 834}$ & $38245 \pm 3242$ & $30707 \pm 4322$ & $\bm{34448 \pm 3016}$ & $21364 \pm 3773$ \\    
  mushroom & $2$ & $18050 \pm 508$ & $\bm{318 \pm 13}$ & $\bm{17930 \pm 506}$ & $681 \pm 32$ & $\bm{17930 \pm 506}$ & $513 \pm 23$  \\    
  connect & $2$ & $15155 \pm 334$ & $\bm{137 \pm 9}$ & $\bm{14978 \pm 340}$ & $228 \pm 17$ & $\bm{14978 \pm 340}$ & $240 \pm 12$   \\    
  connect & $3$ & $179788 \pm 6242$ & $\bm{9897 \pm 602}$ & $64533 \pm 4352$ & $30381 \pm 4267$ & $\bm{58247 \pm 4053}$ & $20106 \pm 2661$  \\    
  munin1 & $2$ & $25180 \pm 266$ & $\bm{699 \pm 20}$ & $\bm{25001 \pm 268}$ & $953 \pm 75$ & $\bm{25001 \pm 268}$ & $1079 \pm 67$  \\    
  andes & $2$ & $29502 \pm 171$ & $\bm{40 \pm 1}$ & $\bm{28810 \pm 182}$ & $249 \pm 37$ & $\bm{28810 \pm 182}$ & $250 \pm 35$ \\    
  andes & $3$ & $92838 \pm 3735$ & $2182 \pm 289$ & $\bm{31162 \pm 136}$ & $504 \pm 132$ & $\bm{31162 \pm 136}$ & $\bm{217 \pm 8}$  \\
  \bottomrule 
  \end{tabular}
  \end{threeparttable}
  \caption{
  Number of binary variables and execution time for the QUBO formulation.  
  The mean and standard deviation of $5$ trials are presented. 
  All the candidate parent subsets not proven optimal within $60$ [s] were feasible solutions. 
  The bold values represent the smallest number of binary variables and the shortest execution times among all approaches. 
  }
  \label{decom1}
\end{table*}

\section{Experimental Results}
\label{Experimental}
In this section, we investigate the performance of our approach for reducing the required number of binary variables. 
We also demonstrate the score maximization using some solvers with our QUBO formulation. 
Version 9.1.2 of the Gurobi Optimizer \footnote{https://www.gurobi.com/} was used to solve the integer linear programming problems. 
The fourth-generation Fujitsu Digital Annealer was employed for score maximization. 
All experiments, except for those using the Digital Annealer, were conducted on a 64-bit Windows machine with an Intel Xeon W-2265 @ $3.50$ GHz and $128$ GB of RAM. 
The code was implemented in the Julia programming language 1.5.3 version. 
As benchmarks, we adopted $5$ simulated datasets from each of the $8$ discrete networks: 
alarm ($n = 37$), barley ($n = 48$), hailfinder ($n = 56$), hepar2 ($n = 70$), win95pts ($n = 76$), pathfinder ($n = 109$), munin1 ($n = 186$), and andes ($n = 223$)
from the Bayesian network repository \footnote{https://www.bnlearn.com/bnrepository/}. 
In addition, we used $5$ times non-recoverable extracted datasets from each of the $3$ datasets: 
chess ($n = 75$), mushroom ($n = 117$), and connect ($n = 129$) from the Frequent Item Set Mining Dataset Repository \footnote{http://fimi.uantwerpen.be/data/}. 
The sample size of each dataset is $1000$.

\subsection{QUBO Formulation} 
We show that our formulation reduces the number of binary variables required in a QUBO. 

\textbf{Settings.~} 
The objective function was $\bar{H} (\bm{u}, \bm{r}, \bm{z})$. 
The score function was the BDeu score \citep{Wray}, where the equivalent sample size was set to $1$. 
See \cref{BDeu}.  
The identification of candidate parent sets was performed exactly. 
Candidate parent sets were identified by simply enumerating all the elements in $\{W \subseteq \mathcal{X} \setminus \{X_i\} \mid |W| \leq m\}$ and comparing their local scores when the inclusion relations between them was in place.  
The maximum parent set size $m$ for each instance with more than $75$ variables was set to the largest value that allowed the identification process to be completed within $48$ [h]. 
The number of candidate parent sets and the execution time for their identification are found in \cref{detail}. 
The integer linear programming problem to find optimal candidate parent subsets was formulated as \cref{Optimization}. 
The time limit for each variable was $60$ [s]. 
The execution time for the QUBO formulation includes that of \cref{SSR}, setting up the integer linear programming problem, solving it, and incorporating the solution into the objective function. 

\textbf{Baselines.~} 
We compare our formulation with the approach proposed by \citep{Bryan}. 
Each vertex requires $\lfloor \frac{1}{4} (|\mathcal{X}_i| - 1)^2 \rfloor + 2$ auxiliary variables when $m_i = 3$, and $\frac{1}{2} |\mathcal{X}_i| (|\mathcal{X}_i| - 1) + 3$ when $m_i = 4$,  
where $m_i \equiv \max_{0 \leq j \leq \lambda_i} |W_{ij}|$. 
To investigate the merit of \cref{SSR}, we found candidate parent subsets using the integer linear programming problem with $W_\mathrm{open} = \mathcal{F}_i, V_\mathrm{open} = \bigcup_{V \in \mathcal{F}_i} 2^V \setminus \{\emptyset\}$, and $V_\mathrm{close} = \emptyset$. 
We refer to this as "IP only" and our approach with \cref{SSR} as "IP + \cref{SSR}". 
The candidate parent sets to formulate these baselines were the same as those in the previously described setting. 
The execution time for each baseline does not include the time spent on identifying candidate parent sets.  

\textbf{Evaluation.~} 
The comparison between IP + \cref{SSR} and IP only in \cref{decom1} demonstrates that optimizing candidate parent subsets with \cref{SSR} is advantageous in terms of both the number of binary variables and execution time when the parent set size is larger.  
The quadratization for some instances with $m = 3$ required binary variables over the $100$K bit capacity of the Digital Annealer. 
The decomposed quadratization in our approach significantly reduced the required number of binary variables compared to the quadratization approach. 
The execution time of our formulation was disadvantageous for most larger networks, primarily due to the increasing number of terms.  
\Cref{decom20} shows that incorporating the candidate parent subsets into the QUBO introduced a significant bottleneck.

\subsection{Score Maximization}
We demonstrate score maximization using the QUBO formulation by IP + \cref{SSR}. 

\textbf{Settings.~} 
The time limit for each trial was set to $3600$ [s]. 
The coefficients in the penalty terms were $\delta_1 = 1.1 \delta_0, \delta_2 = 1.1 (n - 2) \delta_1$, and $\xi = 1.1 \xi_0$. 
We used two heuristic solvers and one exact method: 
the Digital Annealer (DAQ), classical simulated annealing (SAQ), and the Gurobi Optimizer (GOQ). 
The inequality constraint is placed outside the objective function in both DAQ and GOQ. 
This formulation is strictly categorized as binary quadratic programming. 
See \cref{BQP}. 
The parameters for DAQ and GOQ were set to their default values. 
The annealing schedule of SAQ was geometric, starting with $T_0 = 100$. 
See \cref{SAQ}.  
Additional bits for a minor embedding \citep{Vicky, Vicky2, Eppstein} were not required because the Digital Annealer is a fully-coupled type. 
Note that the short description of minor embedding can be found in \cref{ME}.  

\textbf{Baselines.~}  
The baseline algorithms are only score-based. 
We adopted three approximate approaches and one exact algorithm: 
the hill climbing search method (HCS), the simulated annealing over ordering space (SAO), the acyclic selection ordering-based search (ASO), and the GOBNILP software \footnote{https://www.cs.york.ac.uk/aig/sw/gobnilp/} (GOB). 
The HCS and SAO are approximate approaches described in \citep{Marco}. 
The ASO and GOB are known as competitive algorithms. 
The tabu list length in HCS was set to $10$, and the state of $10$ randomly selected possible edges was changed upon termination of the greedy search. 
The annealing schedule of SAO was same as that of SAQ. 
The ASO was the version proposed in \citep{Mauro}. 
The version of GOB was the pygobnilp1.0 that relies on the Gurobi Optimizer. 
We used common candidate parent sets in our formulation, except for HCS. 
To ensure a fair comparison, the time limit for each of SAO, ASO, and GOB was set at $3600$ [s] plus the execution time required for the QUBO formulation by IP + \cref{SSR}. 
The time limit for HCS was further extended by the time needed to identify candidate parent sets. 
We ignored the time to transfer the encoding information online to the Digital Annealer environment. 
The HCS and ASO were restarted repeatedly during the time limit. 

\textbf{Evaluation.~} 
\Cref{decom3} displays the results of score maximization. 
For the smaller instances, GOB identified optimal networks. 
In most other cases, the scores achieved by DAQ exceeded those of the baselines. 
If the long execution time required for the QUBO formulation can be reduced in future work, the difference will increase further. 
The Digital Annealer effectively highlighted the strengths of our formulation compared to the results from SAQ and GOQ using the same QUBO formulation.

\begin{table}[t] 
  \centering
  \begin{threeparttable}[]
  \begin{tabular}{>{\centering}p{1.3cm} >{\centering}p{0.3cm} p{5.5cm}} 
  \toprule
  Instance & $m$ & \makecell{\centering Time [s]}    \\ 
  \midrule 
  alarm & $4$ & $1 \pm 0, 3 \pm 1, 10 \pm 1$   \\
  barley & $4$ & $0 \pm 0, 0 \pm 0, 11 \pm 2$   \\
  hailfinder & $4$ & $0 \pm 0, 0 \pm 0, 4 \pm 1$   \\
  hepar2 & $4$ & $0 \pm 0, 0 \pm 0, 2 \pm 0$   \\
  \midrule 
  chess & $2$ & $2 \pm 0, 0 \pm 0, 100 \pm 4$   \\
  chess & $3$ & $406 \pm 49, 2554 \pm 199, 2608 \pm 195$   \\
  win95pts & $2$ & $1 \pm 0, 0 \pm 0, 36 \pm 4$   \\
  win95pts & $3$ & $45 \pm 5, 398 \pm 82, 446 \pm 54$   \\
  pathfinder & $2$ & $3 \pm 0, 0 \pm 0, 281 \pm 19$   \\
  pathfinder & $3$ & $1407 \pm 272, 1679 \pm 184, 18278 \pm 3349$   \\
  mushroom & $2$ & $23 \pm 1, 0 \pm 0, 490 \pm 22$   \\
  connect & $2$ & $5 \pm 0, 0 \pm 0, 235 \pm 12$   \\
  connect & $3$ & $2472 \pm 436, 4129 \pm 148, 13505 \pm 2114$   \\
  munin1 & $2$ & $12 \pm 1, 0 \pm 0, 1067 \pm 67$   \\
  andes & $2$ & $1 \pm 0, 0 \pm 0, 249 \pm 35$   \\  
  andes & $3$ & $6 \pm 1, 2 \pm 1, 209 \pm 8$   \\
  \bottomrule 
  \end{tabular}
  \end{threeparttable}
  \caption{
  Execution time for IP + \cref{SSR}. 
  The execution time for each of the following three steps is displayed. 
  Left : \cref{SSR} and preparation of the interger linear programming problem. 
  Middle : finding candidate parent subsets by solving the problem. 
  Right : incorporating the candidate parent subsets into the objective function of the QUBO. 
  Note that the execution time for IP only is shown in \cref{detail2}. 
  }
  \label{decom20}
\end{table}

\begin{table}[t]
    \centering
    \begin{threeparttable}[]
    \begin{tabular}{ccccccc} 
    \toprule
    Instance & $m$ & DAQ & GOQ & HCS & SAO & GOB    \\ 
    \midrule 
    alarm & $4$ & $0$ & $0$ & $0$ & $0$ & $\bm{5}$ \\    
    barley & $4$ & $0$ & $0$ & $0$  & $0$ & $\bm{5}$   \\    
    hailfinder & $4$ & $0$ & $0$ & $0$ & $0$ & $\bm{5}$   \\    
    hepar2 & $4$ & $0$ & $0$ & $0$  & $0$ & $\bm{5}$   \\    
    \midrule   
    chess & $2$ & $\bm{5}$ & $0$ & $0$ & $0$ & $0$ \\    
    chess & $3$ & $\bm{5}$ & $0$ & $0$  & $0$ & $0$   \\    
    win95pts & $2$ & $1$ & $0$ & $0$ & $0$ & $\bm{4}$   \\    
    win95pts & $3$ & $2$ & $0$ & $0$ & $\bm{3}$ & $0$   \\    
    pathfinder & $2$ & $\bm{4}$ & $0$ & $0$ & $1$ & $0$   \\    
    pathfinder & $3$ & $\bm{5}$ & $0$ & $0$ & $0$ & $0$   \\    
    mushroom & $2$ & $\bm{5}$ & $0$ & $0$ & $0$ & $0$   \\    
    connect & $2$ & $\bm{5}$ & $0$ & $0$ & $0$ & $0$   \\    
    connect & $3$ & $\bm{2}$ & $1$ & $1$ & $1$ & $0$   \\    
    munin1 & $2$ & $\bm{5}$ & $0$ & $0$ & $0$ & $0$   \\    
    andes & $2$ & $\bm{5}$ & $0$ & $0$ & $0$ & $0$   \\
    andes & $3$ & $\bm{5}$ & $0$ & $0$ & $0$ & $0$   \\
    \bottomrule 
    \end{tabular}
    \end{threeparttable}
    \caption{
    Results of score maximization. 
    The number of trials achieving the highest score is displayed. 
    The highest score refers to the shortest execution time when the scores are the same.   
    For the four instances above the line, the GOB identified optimal networks within a few seconds. 
    While the GOQ identified an optimal network in only one trial of the barley instance, the execution time was longer than the GOB.    
    The solutions for the win95pts with $m = 2$ and the connect with $m = 3$ were not proven to be optimal. 
    The SAQ and ASO never won in any of the trials. 
    The bold values show the greatest number of wins among all solvers. 
    }
    \label{decom3}
\end{table}

\section{Conclusion} 
\label{conclusion}
We proposed a QUBO formulation tailored to score-based Bayesian network structure learning. 
The essence of this approach lies in reducing the number of required binary variables through decomposed quadratization with candidate parent subsets.  
We also provided an algorithm to efficiently find optimal candidate parent subsets. 
Experimental results demonstrated that our approach significantly reduced the number of binary variables compared to the previous work based on quadratization. 
Additionally, our formulation using the Digital Annealer achieved improved BDeu scores over existing methods for medium-sized instances. 
We expect that our approach can be more effectively applied to larger-scale structure learning problems in the future development of annealing processors. 

\textbf{Limitations.~} 
We have identified three limitations in our approach. 
First, the execution time to incorporate the candidate parent subsets into the objective function is a potential drawback for larger-scale problems.  
Secondly, capturing a parent set with two bits may be disadvantageous for a 1-bit inversion search, as transitioning from one candidate parent set to another necessitates an inversion of up to four bits.
Lastly, our formulation could potentially have a negative impact by increasing the number of terms in a QUBO. 
In particular, a minor embedding requires additional bits as the number of terms increases. 

\textbf{Future Works.~} 
Along with further investigation addressing the above limitations, we will explore the application of decomposed quadratization to tasks beyond Bayesian network structure learning. 
Decomposed quadratization can serve as a promising alternative to conventional quadratization for a wide range of tasks involving multilinear polynomials with higher-degree terms.

\section*{Acknowledgments} 
We acknowledge Fujitsu Limited for their support in providing access to the fourth-generation Fujitsu Digital Annealer.

\bigskip
\bibliography{AAAI2025_DA}

\newpage 
\appendix
\onecolumn

\section{Notations}
\label{Description}
\begin{table*}[h]
  \centering
  \begin{threeparttable}[]
  \begin{tabular}{c|l} 
    \toprule
    Notation  & \multicolumn{1}{c}{Description}  \\ 	
    \midrule 
    $n$ & the number of vertices  \\ 
    $m$ & the maximum parent set size  \\ 
    $\mathcal{X}$ & $\mathcal{X} \equiv \{X_i\}_{i=1}^n$ is the set of vertices  \\ 
    $\bm{d}$ & $\bm{d} \equiv ((d_{ij})_{j = 1}^n)_{i = 1}^n$ with $d_{ij} \in \{0, 1\}$ corresponds to the set of edges   \\  
    $\bm{r}$ & $\bm{r} \equiv (r_{ij})_{(i, j) \in \mathcal{E}}$ with $r_{ij} \in \{0, 1\}$ corresponds to the topological ordering  \\ 
    $\bm{p}$ & $\bm{p} \equiv (p_i)_{i=1}^I$ with $p_i \in \{0, 1\}$ is the set of primary variables  \\ 
    $\bm{q}$ & $\bm{q} \equiv (q_i)_{i=1}^J$ with $q_i \in \{0, 1\}$ is the set of auxiliary variables  \\ 
    $\bm{u}$ & $\bm{u} \equiv ((u_{ij})_{j = 1}^{\mu_i})_{i = 1}^n$ with $u_{ij} \in \{0, 1\}$ is the set of candidate parent subsets \\ 
    $\bm{v}$ & $\bm{v} \equiv (v_i)_{i = 1}^{\nu}$ with $v_i \in \{0, 1\}$ captures the search space of auxiliary variables \\ 
    $\bm{z}$ & $\bm{z} \equiv (z_i)_{i = 1}^{n}$ with $z_i \in \{0, 1\}$ is used to ensure that each vertex has exactly one parent set   \\ 
    $\pi$ & $\pi : 2^{\{1, \cdots, I\}} \rightarrow \mathbb{R} \setminus \{0\}$ returns the coefficients in a multilinear polynomial \\  
    $\bar{\pi}$ & $\bar{\pi} : 2^{\{1, \cdots, I\}} \times 2^{\{1, \cdots, I\}} \rightarrow \mathbb{R}$ returns the coefficients in a decomposed quadratization   \\    
    $\mathcal{F}$ & $\mathcal{F} \subseteq 2^{\{1, \cdots, I\}} \setminus \{\emptyset\}$ is the terms in a multilinear polynomial \\  
    $\mathcal{E}$ & the set of edges on possible cycles given candidate parent sets \\ 
    $Q_i$ & a subset of $2^{\{1, \cdots, I\}} \setminus \{\emptyset\}$ \\ 
    $W_{ij}$ & a candidate parent set of $X_i$ \\ 
    $U_{ij}$ & a candidate parent subset of $X_i$ \\ 
    $\mathcal{F}_i$ & candidate parent sets of $X_i$ except for the empty set \\ 
    $\mathcal{X}_i$ & the union of the candidate parent sets of $X_i$ \\      
    $m_i$ & the maximum parent set size of $X_i$    \\ 
    $S_i$ & $S_i : 2^{\mathcal{X} \setminus \{X_i\}} \rightarrow \mathbb{R}$ represents the local score function for a parent set of $X_i$  \\  
    $\Pi_i$ & the set of vertices that have edges directed towards $X_i$ \\ 
    $\pi_i$ & $\pi_i(W) \equiv \sum_{W^\prime \subseteq W} (-1)^{1 + |W \setminus W^\prime|} \log S_i (W^\prime)$ \\   
    $\bar{\pi}_i$ & $\bar{\pi}_i(U, U^\prime) \equiv (-1)^{1 + |\{U\} \cup \{U^\prime\}|} (\log S_i (U \cup U^\prime) - \log S_i (U) - \log S_i (U^\prime) + \log S_i (\emptyset))$ \\      
    $h$ & $h : \{0, 1\}^J \rightarrow [0, \infty)$ is a quadratic polynomial function \\    
    $R$ & $R(r_{ij}, r_{jk}, r_{ik}) \equiv r_{ik} + r_{ij} r_{jk} - r_{ij} r_{ik} - r_{jk} r_{ik}$  \\ 
    $O$ & $O (\bm{d}) \equiv \sum_{1 \leq i \leq n} \sum_{W \subseteq \mathcal{X}_i} \pi_i(W) \prod_{1 \leq j \leq n; X_j \in W} d_{ij}$ \\ 
    $C$ & $C (\bm{d}, \bm{r}) \equiv \sum_{(i, j), (j, k), (i, k) \in \mathcal{E}} \delta_1 R (r_{ij}, r_{jk}, r_{ik}) + \sum_{(i, j) \in \mathcal{E}} \delta_2 \bigl(d_{ij} r_{ij} + d_{ji} (1 -r_{ij})\bigr)$ \\ 
    $\bar{O}$ & $\bar{O}(\bm{u}, \bm{z}) \equiv \sum_{1 \leq i \leq n} \bigl(\sum_{1 \leq j \leq k \leq \mu_i} \bar{\pi}_i (U_{ij}, U_{ik}) u_{ij} u_{ik} + \xi \bigl(z_i - z_i \sum_{1 \leq j \leq \mu_i} u_{ij} + \sum_{1 \leq j < k \leq \mu_i} u_{ij} u_{ik}\bigr)\bigr)$  \\  
    $\bar{C}$ & $\bar{C} (\bm{u}, \bm{r})$ is defined as $C(\bm{d}, \bm{r})$ where $d_{ij}$ is replaced by $\sum_{1 \leq k \leq \mu_i; X_j \in U_{ik}} u_{ik}$  \\   
    $D$ & $D(\bm{u}) \equiv ((\min\{1, \sum_{1 \leq k \leq \mu_i; X_j \in U_{ik}} u_{ik}\})_{j = 1}^n)_{i = 1}^n$ \\   
    $\bar{H}$ & $\bar{H}(\bm{u}, \bm{r}, \bm{z}) \equiv \bar{O}(\bm{u}, \bm{z}) + \bar{C} (\bm{u}, \bm{r})$ \\   
    $\bar{g}_1$ & $\bar{g}_1 (\bm{q}) \equiv (\min \{1, \sum_{1 \leq j \leq J; i \in Q_j} q_j\})_{i=1}^I$  \\ 
    $\bar{g}_2$ & $\bar{g}_2 (\bm{q}) \equiv \sum_{1 \leq i \leq j \leq J} \bar{\pi} (Q_i, Q_j) q_i q_j + h(\bm{q})$  \\ 
    $\xi_0$ & $\xi_0 \equiv - 3 \min_{1 \leq i \leq n} \min_{0 \leq j \leq k \leq \mu_i} \bar{\pi}_i (U_{ij}, U_{ik})$ \\  
    $\delta_0$ & $\delta_0 \equiv \max_{1 \leq i \leq n, 0 \leq j \leq \lambda_i} (\log S_i (W_{ij}) - \log S_i (\emptyset))$  \\   
    $V_\mathrm{close}$ & a subset of an optimal solution \\ 
    $V_\mathrm{open}$ & $V_\mathrm{close} \cup V_\mathrm{open}$ contains an optimal solution \\ 
    $W_\mathrm{open}$ & the terms that still cannot be ignored in $\mathcal{F}$ \\ 
    \bottomrule 
  \end{tabular}
  \end{threeparttable}
  \label{symbol_description}
\end{table*}

\newpage 
\section{Proof} 
\label{proofs}

\subsection{Proof of \Cref{theorem_decomposition}}
\label{proof_dq}
Initially, we introduce the following lemmas. 
\begin{lemma}
  \label{barc0} 
  Let $h_0$ be defined as   
  \begin{align} 
    h_0 (\bm{q}) \equiv \sum_{1 \leq i \leq J} \sum_{\substack{\emptyset \subset \mathcal{W} \subseteq \{1, \cdots, J\} \setminus \{i\} \\ Q_i \subseteq \bigcup_{j \in \mathcal{W}} Q_j \mathrm{~and~} Q_i \nsubseteq \bigcup_{j \in \mathcal{W}^\prime} Q_j \mathrm{~for~any~} \mathcal{W}^\prime \subset \mathcal{W}}} (1 - q_i) \prod_{j \in \mathcal{W}} q_j. \nonumber 
  \end{align} 
  Suppose that $\bm{p} = \bar{g}_1 (\bm{q})$ holds.  
  Then we have 
  \begin{align}
    h_0(\bm{q}) = 0 \Rightarrow \bm{q} = (\prod_{1 \leq i \leq I; i \in Q_j} p_i)_{j=1}^J \mathrm{~~for~any~} \bm{q} \in \{0, 1\}^J. \nonumber 
  \end{align} 
\end{lemma}  
\begin{proof} 
  From $\bm{p} = \bar{g}_1 (\bm{q})$, if $p_i = 1$, then there exists an index $j$ such that $i \in Q_j$ and $q_j = 1$. 
  Therefore, if $h_0(\bm{q}) = 0$, then $Q_j \subseteq \bigcup_{1 \leq i \leq I; p_i = 1} \{i\} \Rightarrow q_j = 1$. 
  Additionally, if $p_i = 0$, then $i \in Q_j \Rightarrow q_j = 0$. 
  Consequently, this lemma holds. 
\end{proof} 
\begin{lemma}
  \label{barc1} 
  Let $h$ be defined as  
  \begin{align} 
    h(\bm{q}) \equiv \kappa \Bigl(\sum_{1 \leq i \leq J} \sum_{\substack{1 \leq j < k \leq J \\ i \notin \{j, k\}, Q_j \cup Q_k = Q_i}} (q_j q_k - 2 q_j q_i - 2 q_k q_i + 3 q_i) + \sum_{\substack{i, j \in \{1, \cdots, J\} \\ Q_i \subset Q_j}} (1 - q_i) q_j\Bigr),  \nonumber 
  \end{align} 
  where $\kappa \in (0, \infty)$.    
  Suppose that \cref{r2} holds. 
  Then we have 
  \begin{align} 
    h(\bm{q}) = 0 \Rightarrow h_0 (\bm{q}) = 0 \mathrm{~~for~any~} \bm{q} \in \{0, 1\}^J.  \nonumber 
  \end{align} 
\end{lemma}  
\begin{proof} 
  We prove $h_0(\bm{q}) \neq 0 \Rightarrow h (\bm{q}) \neq 0$ for any $\bm{q} \in \{0, 1\}^J$. 
  If $h_0(\bm{q}) \neq 0$, there exists at least one pair of an index $i$ and a set $\mathcal{W}$ such that $q_i = 0$ and $\prod_{j \in \mathcal{W}} q_j = 1$ in $h_0(\bm{q})$. 
  For $|\mathcal{W}| = 1$, the second term of $h (\bm{q})$ is positive. 
  Consider the case where $|\mathcal{W}| \geq 2$. 
  From \cref{r2}, there exists $(a, b)$ such that $Q_a \cup Q_b = Q_i$ and $i \notin \{a, b\}$. 
  By repeating this split, we reach $Q_c$ such that $Q_c \subseteq Q_j$ for some $j \in \mathcal{W}$.  
  Therefore, either the first or the second term of $h (\bm{q})$ is positive. 
  Consequently, this lemma holds. 
\end{proof} 
Let $\bar{\pi}$ satisfy 
\begin{align} 
  \sum_{1 \leq i \leq j \leq J; Q_i \cup Q_j = V} \bar{\pi} (Q_i, Q_j) = 
  \left\{ 
    \begin{array}{ll}
      \pi (V) & (V \in \mathcal{F}) \\ 
      0 & (V \notin \mathcal{F}). 
    \end{array}  
  \right. \nonumber     
\end{align}   
From \cref{barc0} and \cref{barc1}, the following holds:  
\begin{align} 
  h (\bm{q}) = 0 \Rightarrow \bm{q} = (\prod_{1 \leq i \leq I; i \in Q_j} p_i)_{j=1}^J \mathrm{~~for~any~} \bm{q} \in \{0, 1\}^J. \nonumber 
\end{align} 
To induce $h (\bm{q}) = 0$, we demonstrate a simple bound. 
Let $\kappa_0$ denote  
\begin{align} 
  \kappa_0 \equiv \max_{\bm{q} \in \{0, 1\}^J} \sum_{1 \leq i \leq j \leq J} \bar{\pi} (Q_i, Q_j) q_i q_j - \min_{\bm{q} \in \{0, 1\}^J} \sum_{1 \leq i \leq j \leq J} \bar{\pi} (Q_i, Q_j) q_i q_j. \nonumber 
\end{align}   
Suppose that $\kappa > \kappa_0$ holds. 
Then the following holds: 
\begin{align} 
  f(\bm{p}) = \min_{\bm{q} \in \{0, 1\}^J; \bm{p} = \bar{g}_1(\bm{q})} \bar{g}_2(\bm{q}). \nonumber 
\end{align}   
Hence, under \cref{r1} and \cref{r2}, $(\bar{g}_1, \bar{g}_2)$ is a decomposed quadratization of $f$.

\subsection{Proof of \Cref{exist0}}  
\label{proof_exist0}
Considering $\phi(1, \cdots, 1, 0) \neq 0$ and ${}_{K} \mathrm{C}_2 + {}_{K} \mathrm{C}_1 + {}_{K} \mathrm{C}_0 < 2^K$, no quadratic polynomial function $\phi : \{0, 1\}^{K + 1} \rightarrow \mathbb{R}$ satisfies \cref{qqq} for $K \in \{3, 4, \cdots\}$.

\subsection{Proof of \Cref{exist1}}  
\label{proof_exist1}
Let $\phi$ be   
\begin{align} 
  \phi(q_1, q_2, q_3) = b_1 q_1 + b_2 q_2 + b_3 q_3 + b_{12} q_1 q_2 + b_{13} q_1 q_3 + b_{23} q_2 q_3. \nonumber 
\end{align}   
From \cref{qqq}, $b_1 = b_2 = 0$ and $b_{12} = \phi(1, 1, 0) \neq 0$. 
For $q_3 = 1$, the following holds:  
\begin{align} 
   &\phi(0, 0, 1) = b_3,&   \nonumber \\ 
   &\phi(1, 0, 1) = b_1 + b_3 + b_{13} = b_3 + b_{13},&   \nonumber \\ 
   &\phi(0, 1, 1) = b_2 + b_3 + b_{23} = b_3 + b_{23},&   \nonumber \\ 
   &\phi(1, 1, 1) = b_1 + b_2 + b_3 + b_{12} + b_{13} + b_{23} = \phi(1, 1, 0) + b_3 + b_{13} + b_{23}.&   \nonumber 
\end{align}   
Consequently, we have  
\begin{align} 
  \phi(0, 0, 1) + \phi(1, 1, 1) - \phi(1, 0, 1) - \phi(0, 1, 1) = \phi(1, 1, 0) \neq 0. \nonumber 
\end{align}

\subsection{Proof of \Cref{sufficient1}}
\label{Proof_Bounds1}
For $u_{ab} = 1$ and $\sum_{1 \leq j \leq \mu_a} u_{aj} > 2$, the following holds: 
\begin{align} 
    \min_{\bm{z} \in \{0, 1\}^n} \bar{O} (\bm{u}, \bm{z}) - \min_{\bm{z} \in \{0, 1\}^n} \bar{O} (\bm{u}^{ab}, \bm{z}) \geq (- 2 + \sum_{1 \leq j \leq \mu_a} u_{aj}) \xi + \sum_{1 \leq j \leq \mu_a} \bar{\pi}_a(U_{ab}, U_{aj}) u_{aj},   \nonumber 
\end{align}   
where $\bm{u}^{ab} \equiv ((u_{ij}^{ab})_{j = 1}^{\mu_i})_{i = 1}^n$ with $u_{ij}^{ab} = 0$ if $(i, j) = (a, b)$; $u_{ij}^{ab} = u_{ij}$ otherwise. 
If $\xi > \xi_0$, then we have 
\begin{align} 
  \min_{\bm{z} \in \{0, 1\}^n} \bar{O} (\bm{u}, \bm{z}) > \min_{\bm{z} \in \{0, 1\}^n} \bar{O} (\bm{u}^{ab}, \bm{z}).   \nonumber 
\end{align}   
This inequality implies that $\sum_{1 \leq j \leq \mu_i} u_{ij} \leq 2$ holds for all $1 \leq i \leq n$. 
Consequently, if $\xi > \xi_0$ holds and the state of $\bm{d}$ aligns with candidate parent sets, then $(D, \bar{O})$ is a decomposed quadratization of $O$.

\subsection{Proof of \Cref{sufficient2}}
\label{Proof_Bounds2}
At first, we introduce the following lemmas. 
\begin{lemma} 
  \label{const_r} 
  Let $T$ be defined as 
  \begin{align} 
    T (\bm{r}) \equiv \sum_{(i, j), (j, k), (i, k) \in \mathcal{E}} R (r_{ij}, r_{jk}, r_{ik}). \nonumber 
  \end{align}   
  Suppose that $T (\bm{r}) > 0$ holds. 
  Then there exists at least one integer pair $(a, b)$ such that $T(\bm{r}) > T(\bm{r}^{ab})$,  
  where $\bm{r}^{ab} \equiv (r_{ij}^{ab})_{(i, j) \in \mathcal{E}}$ with $r_{ij}^{ab} = 1 - r_{ij}$ if $(i, j) = (a, b)$; $r_{ij}^{ab} = r_{ij}$ otherwise. 
\end{lemma} 
\begin{proof} 
  Assume $R(r_{ij}, r_{jk}, r_{ik}) > 0$. 
  For all $l$ satisfying $k < l \leq n$, the following holds: 
  \begin{align} 
    &R(r_{ij}, r_{jl}, r_{il}) - R(1 - r_{ij}, r_{jl}, r_{il}) + R(r_{jk}, r_{kl}, r_{jl}) - R(1 - r_{jk}, r_{kl}, r_{jl}) + R(r_{ik}, r_{kl}, r_{il}) - R(1 - r_{ik}, r_{kl}, r_{il})& \nonumber \\  
    &= (r_{il} + r_{ij} r_{jl} - r_{ij} r_{il} - r_{jl} r_{il}) - (r_{jl} + r_{ij} r_{il} - r_{ij} r_{jl} - r_{jl} r_{il})& \nonumber \\   
    &~~~+ (r_{jl} + r_{jk} r_{kl} - r_{jk} r_{jl} - r_{kl} r_{jl}) - (r_{kl} + r_{jk} r_{jl} - r_{jk} r_{kl} - r_{kl} r_{jl})& \nonumber \\   
    &~~~+ (r_{il} + r_{ik} r_{kl} - r_{ik} r_{il} - r_{kl} r_{il}) - (r_{kl} + r_{ik} r_{il} - r_{ik} r_{kl} - r_{kl} r_{il})& \nonumber \\  
    &= 2 r_{il} (1 - r_{ij} - r_{ik}) + 2 r_{jl} (r_{ij} - r_{jk}) + 2 r_{kl} (r_{jk} + r_{ik} - 1)& \nonumber \\   
    &= 0,& \nonumber  
  \end{align} 
  where $(r_{ij}, r_{jk}, r_{ik})$ is $(1, 1, 0)$ or $(0, 0, 1)$.   
  For all $l$ satisfying $j < l < k$, the following holds: 
  \begin{align} 
    &R(r_{ij}, r_{jl}, r_{il}) - R(1 - r_{ij}, r_{jl}, r_{il}) + R(r_{jl}, r_{lk}, r_{jk}) - R(r_{jl}, r_{lk}, 1 - r_{jk}) + R(r_{il}, r_{lk}, r_{ik}) - R(r_{il}, r_{lk}, 1 - r_{ik})& \nonumber \\  
    &= (r_{il} + r_{ij} r_{jl} - r_{ij} r_{il} - r_{jl} r_{il}) - (r_{jl} + r_{ij} r_{il} - r_{ij} r_{jl} - r_{jl} r_{il})& \nonumber \\   
    &~~~+ (r_{jk} + r_{jl} r_{lk} - r_{jl} r_{jk} - r_{lk} r_{jk}) - (1 - r_{jk} - r_{lk} - r_{jl} + r_{jl} r_{lk} + r_{jl} r_{jk} + r_{lk} r_{jk})& \nonumber \\   
    &~~~+ (r_{ik} + r_{il} r_{lk} - r_{il} r_{ik} - r_{lk} r_{ik}) - (1 - r_{ik} - r_{il} - r_{lk} + r_{il} r_{lk} + r_{il} r_{ik} + r_{lk} r_{ik})& \nonumber \\  
    &= 2 r_{il} (1 - r_{ik} - r_{ij}) + 2 r_{lk} (1 - r_{ik} - r_{jk}) + 2 r_{jl} (r_{ij} - r_{jk}) + 2 (r_{ik} + r_{jk} - 1)& \nonumber \\   
    &= 0.& \nonumber  
  \end{align} 
  For all $l$ satisfying $i < l < j$, the following holds: 
  \begin{align} 
    &R(r_{il}, r_{lj}, r_{ij}) - R(r_{il}, r_{lj}, 1 - r_{ij}) + R(r_{lj}, r_{jk}, r_{lk}) - R(r_{lj}, 1 - r_{jk}, r_{lk}) + R(r_{il}, r_{lk}, r_{ik}) - R(r_{il}, r_{lk}, 1 - r_{ik})& \nonumber \\  
    &= (r_{ij} + r_{il} r_{lj} - r_{il} r_{ij} - r_{lj} r_{ij}) - (1 - r_{ij} - r_{il} - r_{lj} + r_{il} r_{lj} + r_{il} r_{ij} + r_{lj} r_{ij})& \nonumber \\   
    &~~~+ (r_{lk} + r_{lj} r_{jk} - r_{lj} r_{lk} - r_{jk} r_{lk}) - (r_{lj} + r_{jk} r_{lk} - r_{lj} r_{jk} - r_{lj} r_{lk})& \nonumber \\   
    &~~~+ (r_{ik} + r_{il} r_{lk} - r_{il} r_{ik} - r_{lk} r_{ik}) - (1 - r_{ik} - r_{il} - r_{lk} + r_{il} r_{lk} + r_{il} r_{ik} + r_{lk} r_{ik})& \nonumber \\  
    &= 2 r_{il} (1 - r_{ik} - r_{ij}) + 2 r_{lk} (1 - r_{ik} - r_{jk}) + 2 (r_{ij} + r_{ik} - 1)& \nonumber \\   
    &= 0.& \nonumber  
  \end{align} 
  For all $l$ satisfying $1 \leq l < i$, the following holds: 
  \begin{align} 
    &R(r_{li}, r_{ij}, r_{lj}) - R(r_{li}, 1 - r_{ij}, r_{lj}) + R(r_{lj}, r_{jk}, r_{lk}) - R(r_{lj}, 1 - r_{jk}, r_{lk}) + R(r_{li}, r_{ik}, r_{lk}) - R(r_{li}, 1 - r_{ik}, r_{ik})& \nonumber \\  
    &= (r_{lj} + r_{li} r_{ij} - r_{li} r_{lj} - r_{ij} r_{lj}) - (r_{li} + r_{ij} r_{lj} - r_{li} r_{ij} - r_{li} r_{lj})& \nonumber \\   
    &~~~+ (r_{lk} + r_{lj} r_{jk} - r_{lj} r_{lk} - r_{jk} r_{lk}) - (r_{lj} + r_{jk} r_{lk} - r_{lj} r_{jk} - r_{lj} r_{lk})& \nonumber \\   
    &~~~+ (r_{lk} + r_{li} r_{ik} - r_{li} r_{lk} - r_{ik} r_{lk}) - (r_{li} + r_{ik} r_{lk} - r_{li} r_{ik} - r_{li} r_{lk})& \nonumber \\  
    &= 2 r_{lk} (1 - r_{ik} - r_{jk}) + 2 r_{li} (r_{ij} + r_{ik} - 1) + 2 r_{lj} (r_{jk} - r_{ij})& \nonumber \\   
    &= 0.& \nonumber  
  \end{align} 
  From $R(r_{ij}, r_{jk}, r_{ik}) > 0$, the following holds: 
  \begin{align} 
    R(r_{ij}, r_{jk}, r_{ik}) - R(1 - r_{ij}, r_{jk}, r_{ik}) + R(r_{ij}, r_{jk}, r_{ik}) - R(r_{ij}, 1 - r_{jk}, r_{ik}) + R(r_{ij}, r_{jk}, r_{ik}) - R(r_{ij}, r_{jk}, 1 - r_{ik}) = 3. \nonumber  
  \end{align} 
  Using these results, we have   
  \begin{align} 
    T(\bm{r}) - T(\bm{r}^{ij}) + T(\bm{r}) - T(\bm{r}^{jk}) + T(\bm{r}) - T(\bm{r}^{ik}) = 3 > 0.  \nonumber  
  \end{align}     
  Consequently, $T(\bm{r}) > T(\bm{r}^{ab})$ holds for at least one index pair $(a, b) \in \{(i, j), (j, k), (i, k)\}$. 
\end{proof}
Let $\bm{u}^{aef} \equiv ((u_{ij}^{aef})_{j = 1}^{\mu_i})_{i = 1}^n$ satisfy 
\begin{align} 
  u_{ae}^{aef} = u_{af}^{aef} = 1,~~u_{aj}^{aef} = 0 \mathrm{~for~} j \notin \{e, f\}, \mathrm{~and~} u_{ij}^{aef} = u_{ij} \mathrm{~for~} i \neq a. \nonumber  
\end{align}   
Assume the following conditions: 
\begin{align} 
  u_{a0} \in \{0,1\},~~(a, b) \in \mathcal{E},~~0 \leq c < d \leq \mu_a,~~0 \leq e \leq f \leq \mu_a,~~u_{ac} = u_{ad} = 1, \nonumber \\ 
  u_{a0} + \cdots + u_{a\mu_a} = 2,~~X_b \in (U_{ac} \cup U_{ad}) \setminus (U_{ae} \cup U_{af}),~~(U_{ae} \cup U_{af}) \subset (U_{ac} \cup U_{ad}). \nonumber  
\end{align}   
We find the range of penalty coefficients so that the difference in the return value of the objective function is negative when the input state changes to the desired state. 
From the discussion in \cref{Proof_Bounds1}, we consider the case where $\sum_{1 \leq j \leq \mu_i} u_{ij} \leq 2$ for all $1 \leq i \leq n$. 
For $\sum_{1 \leq j \leq \mu_b; X_a \in U_{bj}} u_{bj} \geq 1$, the following holds: 
\begin{align} 
  \min_{\bm{z} \in \{0, 1\}^n} \bar{H}(\bm{u}, \bm{r}, \bm{z}) - \min_{\bm{z} \in \{0, 1\}^n} \bar{H}(\bm{u}^{aef}, \bm{r}, \bm{z}) \geq \delta_2 - \log S_a (U_{ac} \cup U_{ad}) + \log S_a (U_{ae} \cup U_{af}).  \nonumber 
\end{align}   
For $\sum_{1 \leq j \leq \mu_b; X_a \in U_{bj}} u_{bj} = 0$ and $r_{ab} = 1$, the following holds: 
\begin{align} 
  \min_{\bm{z} \in \{0, 1\}^n} \bar{H}(\bm{u}, \bm{r}, \bm{z}) - \min_{\bm{z} \in \{0, 1\}^n} \bar{H}(\bm{u}, \bm{r}^{ab}, \bm{z}) \geq \delta_2 - (n - 2) \delta_1.   \nonumber 
\end{align}   
From \cref{const_r}, we could repeat reversing one element from $\bm{r}$ with decreasing $T(\bm{r})$ until $T(\bm{r}) = 0$. 
For $T(\bm{r}) > T(\bm{r}^{ab})$, $r_{ab} = 0$, and $\sum_{1 \leq j \leq \mu_b; X_a \in U_{bj}} u_{bj} = 0$, the following holds: 
\begin{align} 
  \min_{\bm{z} \in \{0, 1\}^n} \bar{H}(\bm{u}, \bm{r}, \bm{z}) - \min_{\bm{z} \in \{0, 1\}^n} \bar{H}(\bm{u}^{aef}, \bm{r}^{ab}, \bm{z}) \geq \delta_1 - \log S_a (U_{ac} \cup U_{ad}) + \log S_a (U_{ae} \cup U_{af}).   \nonumber 
\end{align}   
Here the following holds: 
\begin{align} 
  \delta_0 \geq \max_{1 \leq a \leq n} \max_{1 \leq b \leq n} \max_{0 \leq c < d \leq \mu_a} \min_{\substack{0 \leq e \leq f \leq \mu_a \\ X_b \in (U_{ac} \cup U_{ad}) \setminus (U_{ae} \cup U_{af}), (U_{ae} \cup U_{af}) \subset (U_{ac} \cup U_{ad})}} \log \frac{S_a (U_{ac} \cup U_{ad})}{S_a (U_{ae} \cup U_{af})}. \nonumber 
\end{align}  
From the above discussion, if $\xi > \xi_0$ and $\delta_2 > (n - 2) \delta_1 > (n - 2) \delta_0$ hold, then the following holds: 
\begin{align} 
  \bar{C} (\bm{u_*}, \bm{r_*}) = 0 \mathrm{~~for~any~} \bar{H}(\bm{u_*}, \bm{r_*}, \bm{z_*}) = \min_{\bm{u}, \bm{r}, \bm{z}} \bar{H}(\bm{u}, \bm{r}, \bm{z}). \nonumber  
\end{align}

\newpage   
\section{Integer Linear Programming for QUBO Formulation} 
\subsection{Formulation for Reducing Binary Variables}
\label{Optimization} 
For any $1 \leq i \leq \nu$, define   
\begin{align}  
\mathcal{W}_i \equiv \{\mathcal{W} \subseteq V_\mathrm{close} \cup V_\mathrm{open} \mid |\mathcal{W}| \geq 2, V_i \subseteq \bigcup_{V \in \mathcal{W}} V, \mathrm{~and~} V_i \nsubseteq \bigcup_{V \in \mathcal{W}^\prime} V \mathrm{~~for~any~} \mathcal{W}^\prime \subset \mathcal{W}\}. \nonumber 
\end{align} 
The search for an optimal solution can be formulated as an integer linear programming problem to minimize $\sum_{1 \leq i \leq \nu} v_i$ under the following constraints:   
\begin{align} 
\sum_{\substack{1 \leq i \leq \nu \\ W \in \bigcup_{V \in \{\emptyset\} \cup V_\mathrm{close}} \{V_i \cup V\}}} v_i + \sum_{\substack{1 \leq i < j \leq \nu \\ W \notin \{V_i, V_j\}, V_i \cup V_j = W}} v_i v_j \geq 1 \mathrm{~~for~all~} W \in W_\mathrm{open}, \nonumber \\ 
v_i \prod_{\substack{1 \leq j \leq \nu \\ V_j \in \mathcal{W}}} v_j \leq \sum_{\substack{V, V^\prime \in V_\mathrm{close} \\ V \cup V^\prime = V_i}} 1 + \sum_{\substack{j \in \{1, \cdots, \nu\} \setminus \{i\} \\ V_i \in \bigcup_{V \in V_\mathrm{close}} \{V_j \cup V\}}} v_j + \sum_{\substack{1 \leq j < k \leq \nu \\ i \notin \{j, k\}, V_j \cup V_k = V_i}} v_j v_k \nonumber \\ 
\mathrm{~~for~all~} 1 \leq i \leq \nu \mathrm{~and~} \mathcal{W} \in \mathcal{W}_i. \nonumber 
\end{align} 
Replacing $v_i v_j$ with $\rho_{ij} \in \{0, 1\}$, we transform a higher degree terms into a linear as  
\begin{align}  
  \sum_{\substack{1 \leq i \leq \nu \\ W \in \bigcup_{V \in \{\emptyset\} \cup V_\mathrm{close}} \{V_i \cup V\}}} v_i + \sum_{\substack{1 \leq i < j \leq \nu \\ W \notin \{V_i, V_j\}, V_i \cup V_j = W}} \rho_{ij} \geq 1 \mathrm{~~for~all~} W \in W_\mathrm{open}, \nonumber  \\ 
  - |\mathcal{W} \setminus V_\mathrm{close}| + v_i + \sum_{\substack{1 \leq j \leq \nu \\ V_j \in \mathcal{W}}} v_j \leq \sum_{\substack{V, V^\prime \in V_\mathrm{close} \\ V \cup V^\prime = V_i}} 1 + \sum_{\substack{j \in \{1, \cdots, \nu\} \setminus \{i\} \\ V_i \in \bigcup_{V \in V_\mathrm{close}} \{V_j \cup V\}}} v_j + \sum_{\substack{1 \leq j < k \leq \nu \\ i \notin \{j, k\}, V_j \cup V_k = V_i}} \rho_{jk} \nonumber \\  
  \mathrm{~for~all~} 1 \leq i \leq \nu \mathrm{~and~} \mathcal{W} \in \mathcal{W}_i, \nonumber \\  
  2 \rho_{ij} \leq v_i + v_j \leq 1 + \rho_{ij} \mathrm{~for~all~} 1 \leq i < j \leq \nu.  \nonumber 
\end{align} 
To find optimal candidate parent subsets, we ignore the latter constraint.

\subsection{Formulation for Term Reduction} 
\label{term}
Consider the formulation shown in \cref{proof_dq}. 
Let $\eta_{ij} \in \{0,1\}$ represent the existence of a term $v_i v_j$. 
The number of terms is reduced by solving an integer programming problem that minimizes $\sum_{1 \leq i \leq j \leq \nu + \nu^\prime} \eta_{ij}$ subject to the following constraints:   
\begin{align} 
  \eta_{ij} \leq v_i v_j  \mathrm{~~for~all~} 1 \leq i \leq j \leq \nu + \nu^\prime,     \nonumber \\ 
  v_i = 1  \mathrm{~~for~all~} \nu + 1 \leq i \leq \nu + \nu^\prime,     \nonumber \\ 
  \eta_{ij} = 1 \mathrm{~~for~all~} \nu + 1 \leq i \leq j \leq \nu + \nu^\prime \mathrm{~and~} V_i \cup V_j \in \mathcal{F},     \nonumber \\ 
  v_i v_j \leq \frac{1}{2} (\eta_{jj} + \eta_{\min\{i,j\}\max\{i,j\}}) \mathrm{~~for~all~} i, j \in \{1, \cdots, \nu + \nu^\prime\} \mathrm{~and~} V_i \subset V_j,     \nonumber \\ 
  - 2 + v_i + v_j + v_k \leq \frac{1}{4} (\eta_{ii} + \eta_{\min\{i,j\}\max\{i,j\}} + \eta_{\min\{i,k\}\max\{i,k\}} + \eta_{jk}) \nonumber \\  
  \mathrm{~~for~all~} 1 \leq i \leq \nu + \nu^\prime, 1 \leq j < k \leq \nu + \nu^\prime, i \notin \{j, k\}, \mathrm{~and~} V_j \cup V_k = V_i, \nonumber \\ 
  \sum_{\substack{1 \leq i \leq j \leq \nu + \nu^\prime \\ V_i \cup V_j = W}} \eta_{ij} \geq 1 \mathrm{~~for~all~} W \in \mathcal{F},  \nonumber \\ 
  - |\mathcal{W}| + v_i + \sum_{\substack{1 \leq j \leq \nu + \nu^\prime \\ V_j \in \mathcal{W}}} v_j \leq \sum_{\substack{1 \leq j < k \leq \nu + \nu^\prime \\ i \notin \{j, k\}, V_j \cup V_k = V_i}} v_j v_k \mathrm{~~for~all~} 1 \leq i \leq \nu \mathrm{~and~} \mathcal{W} \in \mathcal{W}_i, \nonumber 
\end{align} 
where $(v_i)_{i = \nu + 1}^{\nu + \nu^\prime} \in \{0,1\}^{\nu^\prime}$ and $\{V_i\}_{i = \nu + 1}^{\nu + \nu^\prime} \equiv V_\mathrm{close}$.   
As described in \cref{Optimization}, we transform quadratic terms into linear ones by replacing $v_i v_j$ with $\rho_{ij}$.

\newpage   
\section{Experiment}

\subsection{Bayesian Dirichlet Equivalence Uniform Score} 
\label{BDeu}
For the Bayesian Dirichlet equivalence uniform (BDeu) score, define 
\begin{align}  
  S_i (\Pi) \equiv \prod_{1 \leq j \leq \beta_i} \frac{\Gamma(\alpha_{ij})}{\Gamma(N_{ij} + \alpha_{ij})} \prod_{1 \leq k \leq \gamma_i} \frac{\Gamma(N_{ijk} + \alpha_{ijk})}{\Gamma(\alpha_{ijk})} \mathrm{~~for~any~} 1 \leq i \leq n \mathrm{~and~} |\Pi| \leq m_i, \nonumber 
\end{align} 
where $N_{ij} \equiv \sum_{1 \leq k \leq \gamma_i} N_{ijk}$, $N_{ijk}$ is the number of cases of $\Pi$ in its $j$-th state and $X_i$ in its $k$-th state, 
$\alpha_{ij} \equiv \sum_{1 \leq k \leq \gamma_i} \alpha_{ijk}$, $\alpha_{ijk} \equiv \frac{\alpha}{\beta_i \gamma_i}$, $\beta_i$ is the number of joint states of $\Pi$, $\gamma_i$ is the number of states of $X_i$, 
and $\alpha \in (0, \infty)$ is the equivalent sample size.

\subsection{Identification of Candidate Parent Sets}
\label{detail} 

\begin{table}[h]
  \centering
  \begin{threeparttable}[]
  \begin{tabular}{cccc} 
  \toprule
  Instance & $m$ & Number & Time [s]  \\ 
  \midrule 
  alarm & $4$ & $2291 \pm 210$ & $1700 \pm 11$   \\ 
  barley & $4$ & $355 \pm 12$ & $235015 \pm 95243$   \\ 
  hailfinder & $4$ & $692 \pm 16$ & $90247 \pm 3334$   \\ 
  hepar2 & $4$ & $689 \pm 59$ & $40216 \pm 534$   \\ 
  \midrule 
  chess & $2$ & $32570 \pm 1686$ & $59 \pm 0$   \\ 
  chess & $3$ & $236285 \pm 15636$ & $5463 \pm 284$   \\ 
  win95pts & $2$ & $9231 \pm 665$ & $60 \pm 3$   \\ 
  win95pts & $3$ & $43657 \pm 3850$ & $2404 \pm 51$   \\ 
  pathfinder & $2$ & $50962 \pm 3254$ & $320 \pm 5$   \\ 
  pathfinder & $3$ & $398131 \pm 47889$ & $44991 \pm 3426$   \\ 
  mushroom & $2$ & $273526 \pm 12647$ & $689 \pm 26$   \\ 
  connect & $2$ & $76996 \pm 4381$ & $331 \pm 7$   \\ 
  connect & $3$ & $710684 \pm 73964$ & $83361 \pm 6624$   \\ 
  munin1 & $2$ & $135570 \pm 6765$ & $2466 \pm 29$   \\ 
  andes & $2$ & $7424 \pm 219$ & $852 \pm 52$   \\ 
  andes & $3$ & $14195 \pm 556$ & $104683 \pm 3855$   \\ 
  \bottomrule 
  \end{tabular}
  \end{threeparttable}
  \caption{
  Number of candidate parent sets and execution time for their identification. 
  The mean and standard deviation of $5$ trials are presented. 
  Empty sets were excluded from the candidate parent sets. 
  }
  \label{CPSI}
\end{table}

\subsection{Classical Simulated Annealing} 
\label{SAQ}
\begin{algorithm*}[h]
  \caption{Classical Simulated Annealing}  
  \begin{algorithmic}[1]
  \WHILE{the current time is within the time limit,}  
    \STATE $\Delta \leftarrow$ the incremental return value of $\bar{H}$ when filipping a bit randomly.  
    \IF{a random number between $0$ and $1$ is less than $\exp (- \Delta / T_0^{1 - \mathrm{current~time} / \mathrm{time~limit}})$,} 
      \STATE accept the bit flip. 
    \ENDIF 
  \ENDWHILE 
  \end{algorithmic}
\end{algorithm*}

\subsection{Binary Quadratic Programming} 
\label{BQP} 
The binary quadratic programming problem for learning a Bayesian network can be defined by 
\begin{align}  
  \mathrm{minimize~} \sum_{1 \leq i \leq n} \sum_{1 \leq j \leq k \leq \mu_i} \bar{\pi}_i (U_{ij}, U_{ik}) u_{ij} u_{ik} + \bar{C} (\bm{u}, \bm{r}) \mathrm{~~~~subject~to~} \sum_{1 \leq j \leq \mu_i} u_{ij} \leq 2 \mathrm{~~for~all~} 1 \leq i \leq n. \nonumber
\end{align} 
Note that we can express the DAG constraint using linear inequalities. 
The linear ordering of vertices is 
\begin{align}  
  0 \leq r_{ij} + r_{jk} - r_{ik} \leq 1 \mathrm{~~for~all~} (i,j), (j,k), (i,k) \in \mathcal{E}.     \nonumber  
\end{align} 
The consistency of edges and ordering is
\begin{align}  
  \sum_{\substack{1 \leq k \leq \mu_i \\ X_j \in U_k}} u_{ik} \leq 2 - 2 r_{ij} \mathrm{~~and~~} \sum_{\substack{1 \leq k \leq \mu_j \\ X_i \in U_k}} u_{jk} \leq 2 r_{ij} \mathrm{~~~for~all~} (i,j) \in \mathcal{E}.     \nonumber  
\end{align}

\subsection{Identification of Candidate Parent Subsets}
\label{detail2} 

\begin{table}[h]
  \centering
  \begin{threeparttable}[]
  \begin{tabular}{>{\centering}p{1.3cm} >{\centering}p{0.3cm} p{5.5cm}} 
  \toprule
  Instance & $m$ & \makecell{\centering Time [s]}    \\ 
  \midrule 
  alarm & $4$ & $1 \pm 0, 50 \pm 28, 10 \pm 1$    \\
  barley & $4$ & $0 \pm 0, 1 \pm 0, 10 \pm 2$    \\
  hailfinder & $4$ & $0 \pm 0, 0 \pm 0, 4 \pm 1$   \\
  hepar2 & $4$ & $0 \pm 0, 0 \pm 0, 2 \pm 0$    \\
  \midrule 
  chess & $2$ & $6 \pm 0, 0 \pm 0, 83 \pm 3$    \\
  chess & $3$ & $639 \pm 58, 3150 \pm 167, 3480 \pm 315$   \\
  win95pts & $2$ & $2 \pm 2, 0 \pm 0, 31 \pm 1$    \\
  win95pts & $3$ & $49 \pm 8, 530 \pm 73, 457 \pm 66$    \\
  pathfinder & $2$ & $18 \pm 2, 0 \pm 0, 241 \pm 29$   \\
  pathfinder & $3$ & $3844 \pm 750, 2132 \pm 129, 24731 \pm 3458$   \\
  mushroom & $2$ & $277 \pm 19, 1 \pm 0, 403 \pm 14$    \\
  connect & $2$ & $26 \pm 3, 0 \pm 0, 203 \pm 15$   \\
  connect & $3$ & $6219 \pm 1396, 4568 \pm 177, 19595 \pm 2743$   \\
  munin1 & $2$ & $78 \pm 6, 0 \pm 0, 875 \pm 71$   \\
  andes & $2$ & $1 \pm 0, 0 \pm 0, 247 \pm 37$    \\  
  andes & $3$ & $4 \pm 0, 7 \pm 1, 493 \pm 131$   \\
  \bottomrule 
  \end{tabular}
  \end{threeparttable}
  \caption{
  Execution time for IP only. 
  Displays are the same as \cref{decom20}. 
  }
\end{table}

\subsection{Minor Embedding} 
\label{ME}
Annealing machines are classified into two types: nearest-neighbor and fully-coupled. 
A fully-coupled annealing machine allows coupling between arbitrary vertices, whereas coupling in a nearest-neighbor annealing machine is limited to adjacent vertices only. 
Nearest-neighbor annealing machines require additional bits to embed a highly dense logical topology into a physical one \citep{Vicky, Vicky2}. 
The number of additional bits for the embedding depends on the design of the hardware graphs, and embedding problems are NP-hard \citep{Eppstein}.

\end{document}